\documentclass[]{article}

\usepackage{graphicx}

\usepackage{framed,multirow}
\usepackage{authblk}

\usepackage{url}
\usepackage{xcolor}
\definecolor{newcolor}{rgb}{.8,.349,.1}

\usepackage[lined,boxed,commentsnumbered]{algorithm2e}
\usepackage[font=small,labelfont=bf,textfont=bf, skip=7pt]{caption}

\usepackage{color}

\newcommand\blfootnote[1]{%
	\begingroup
	\renewcommand\thefootnote{}\footnote{#1}%
	\addtocounter{footnote}{-1}%
	\endgroup
}

\title{Handwritten Isolated Bangla Compound Character Recognition: a new benchmark using a novel deep learning approach}
\author[1]{Saikat Roy}
\author[2]{Nibaran Das} 
\author[2]{Mahantapas Kundu}
\author[2]{Mita Nasipuri}

\affil[1]{Dept. of Information Technology, Jadavpur University, India}
\affil[2]{Dept. of Computer Science \& Engg., Jadavpur University, India}

\date{}
\begin{document}

\maketitle

\begin{abstract}
In this work, a novel deep learning technique for the recognition of handwritten \textit{Bangla} isolated compound character is presented and a new benchmark of recognition accuracy on the CMATERdb 3.1.3.3 dataset is reported. Greedy layer wise training of Deep Neural Network has helped to make significant strides in various pattern recognition problems. We employ \textit{layerwise training} to Deep Convolutional Neural Networks (DCNN) in a supervised fashion and augment the training process with the RMSProp algorithm to achieve faster convergence. We compare results with those obtained from standard shallow learning methods with predefined features, as well as standard DCNNs. Supervised layerwise trained DCNNs are found to outperform standard shallow learning models such as Support Vector Machines as well as regular DCNNs of similar architecture by achieving  error rate of 9.67\% thereby setting a new benchmark on the CMATERdb 3.1.3.3 with recognition accuracy of 90.33\%, representing an improvement of nearly 10\%.
\end{abstract}

\section{Introduction}

\blfootnote{Preprint, since published in \textit{Pattern Recognition Letters, Elsevier} (Recommended)} 
\blfootnote {DOI: https://doi.org/10.1016/j.patrec.2017.03.004}

Deep learning maybe loosely defined as an attempt to train a hierarchy of feature detectors -- with each layer learning a higher representation of the preceding layer. The advent of deep learning has seen a resurgence in the use of (increasingly larger) neural networks \cite{Szegedy2014}. Among many other areas of application of deep neural networks (DNNs), handwritten character recognition is one of the areas that has been extensively explored, particularly on the MNIST dataset \cite{LeCun1998}.

\begin{figure}[!t]
	\centering
	\includegraphics[scale=.35]{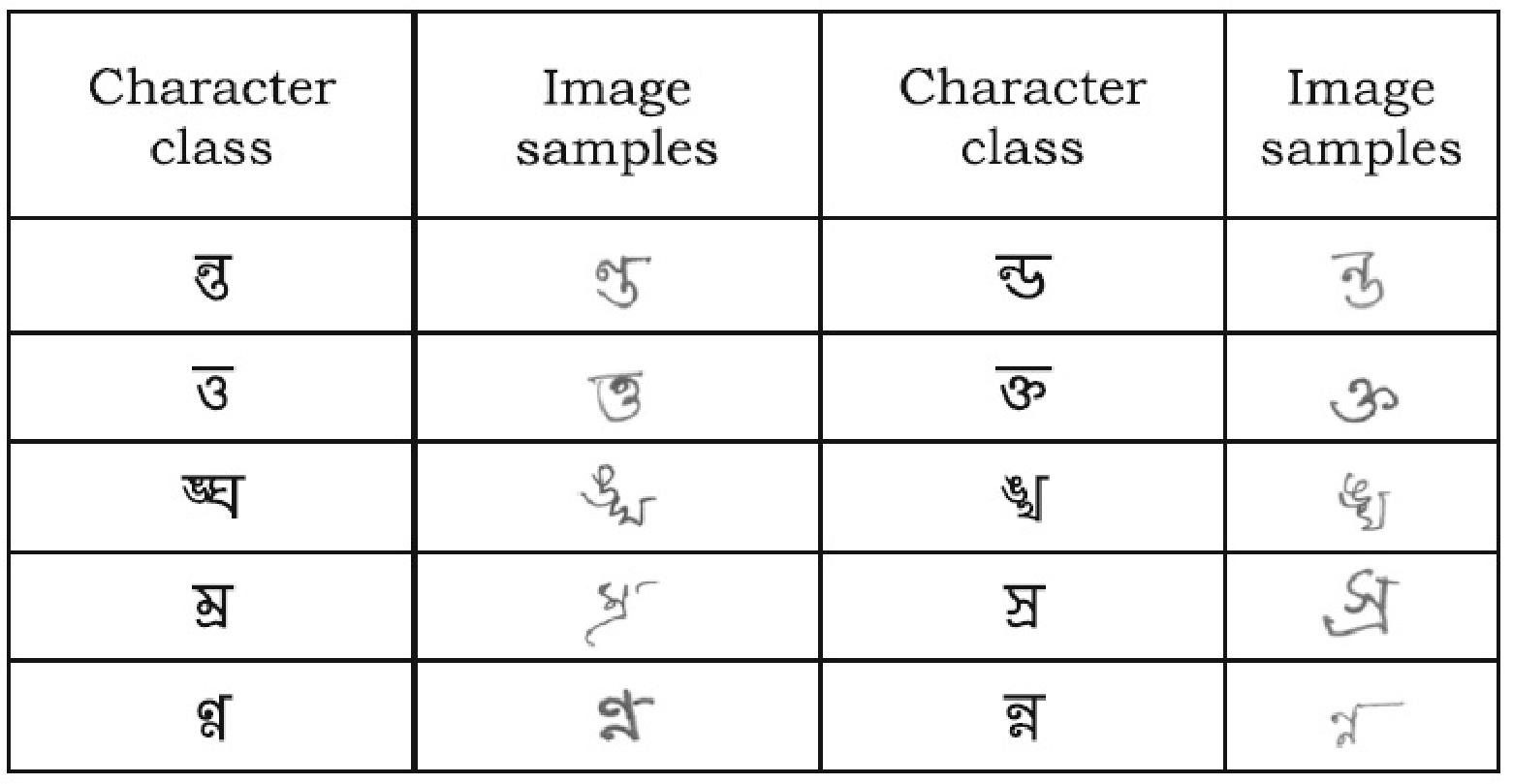}
	\caption{\protect Random samples from a few closely resembling classes \cite{Das2014}}
	\label{fig:hardproblem}
\end{figure}

But very few works have been published for Indian languages specially for \textit{Bangla} which is sixth most popular language in the world \cite{bengali_wiki}. \textit{Bangla} language consists of 11 vowels, 39 consonants, 10 modifiers and 334 compound characters. Among them, compound characters are structurally complex and some of them resemble so closely Figure \ref{fig:hardproblem}) that the only sign of differences left between them are short straight lines, circular curves etc. \cite{Das2014} Due to this and the high number of classes, compound character recognition is a particularly challenging pattern recognition problem .

On the other hand, Deep learning has shown promise in recent years in multiple varied fields including handwritten character, speech and object recognition \cite{Krizhevsky2012,Hinton2012,NIPS2009_1171,Goodfellow2013}, natural language processing \cite{He2014} etc. Owing to its successful application in above areas, it is applied on the problem of handwritten \textit{Bangla} compound character recognition which is much more challenging than MNIST dataset which consists of only 10 classes.

Deep learning methods usually benefit from a substantial amount of labeled or unlabeled data to build a powerful model to represent the data. The \textit{Bangla} compound character datasets used in this work (Section \ref{sec:dataset}), is freely downloadable for OCR research and consists of 171 unique classes with approximately 200 samples per class. The amount of data available, hence, necessitates the use of easily generalizable deep learning models or techniques to synthetically augment the data while training. Although shallow learning models, which have been previously used for \textit{Bangla} compound character recognition, benefit from not having such requirements, the lower recognition accuracy produced by such models offset the benefits of not requiring a large amount of data.

Shallow learning models that are still popular for various tasks take an approach to pattern recognition which prioritizes the usage of \textit{feature engineering} in extracting unique, robust and discriminating features. These features are then fed to machine learning models for classification or regression purposes. Over the last few decades various powerful multipurpose feature detectors have gained popularity such as HOG \cite{Dalal1467360}, SIFT \cite{SIFT2004,Surinta2015405}, LBP \cite{Ojala199651}, SURF \cite{surf2006} etc. which have all been developed for a particular problem. But the major problem with such methodologies is that the usage of feature detectors need to be guided by the skill of the researchers using them.

\begin{table}[t]
	
	\setlength{\tabcolsep}{8pt}
	\caption{Major differences between MNIST and CMATERdb 3.1.3.3}
	\begin{center}
		\begin{tabular}{|| c || c | c ||} 
			\hline
			\textbf{Parameter} & \textbf{MNIST} & \textbf{CMATERdb 3.1.3.3} \\ [0.5ex]
			\hline\hline
			Scale & Uniform & Non-uniform \\ 
			\hline
			Translation & Centred & Non-Centred \\
			\hline
			Problem Size & $10$ class & $171$ class \\
			\hline
			Training Data & $60000$ samples & $34439$ samples \\
			\hline
			Testing Data & $10000$ samples & $8520$ samples \\ [1ex] 
			\hline
		\end{tabular}
	\end{center}
	\label{table:diffmnistcmater}
	
\end{table}

Deep learning subverts the usage of handcrafted features by learning the features which are most effective for a particular problem. Inspired by Hubel and Wiesel's early work on the cat's visual cortex \cite{HUBEL1959,HUBEL1965}, the convolutional neural network improved on the Neocognitron model \cite{Fukushima1980,Fukushima1983}. It was subsequently used in areas of handwriting recognition and more in the 80s and 90s \cite{LeCun1989,LeCun1995,LeCunJackelB.BoserJ.S.DenkerD.HendersonR.E.HowardW.Hubbard1990} and showed promise for being used in deeper feedforward architectures. However, it was restricted by the limitations of computer hardware of the time. The LeNet model represented a culmination of sorts of work on CNNs in the 90s \cite{LeCun1998}.

Although neural network models deeper than 4 layers had not been typically used till the mid of the first decade of the 2000s, the convolutional neural network had already been established as viable machine learning architecture \cite{Simard2003,LeCun1998}. Subsequent years saw major works being done in the field of deep learning, particularly increasing usage of unsupervised training \cite{Hinton2006} and the greedy unsupervised layerwise training of DNNs \cite{Bengio2007}. The use of stacked denoising autoencoders \cite{Vincent2008,Vincent2010}, rectified linear activation units \cite{Krizhevsky2012,he2015delving} -- which were resistant to the vanishing gradient problem \cite{Hochreiter1998a} -- and the introduction of GPGPU Programming pushed the \textit{depth} of neural networks \cite{Szegedy2014} even further.

During the last decade, the deep convolutional neural networks have carved out a large role in the discussion on deep learning. The popularization of Max Pooling \cite{Scherer2010}, introduction of Dropout \cite{Srivastava2013}, Maxout networks \cite{Goodfellow2013,cai2016maxout} etc. led to widespread applications of DCNNs and DNNs in general. They generally encompassed character recognition, object detection, speech recognition, medical imaging amonng other pattern recognition problems \cite{Ciresan2010,Ciresan2013,Goodfellow2013,shin2016deep,havaei2016brain,jaderberg2016reading,abdel2016breast,ballester2016performance}. Various datasets such as MNIST, TIMIT, ImageNet, CIFAR and SVHN saw new benchmarks set on them using DNNs. With the availability of parallel programming on GPUs \cite{Krizhevsky2012,Ciresan2011,li2016optimizing}, interest in the development of architectures deeper than 15 layers \cite{VGG1619,sercu2016very} and transferring the experience of models trained for one domain to another \cite{holder2016road,shouno2015transfer}, the explosion of interest in deep learning research is a well established fact. 

\section{Previous work}

\subsection{Literature of Handwritten \textit{Bangla} Compound Character Recognition}
Ample works on handwritten \textit{Bangla} character recognition have been reported through the last decade. There are various research works which have attempted to deal with the history of \textit{Bangla} character recognition \cite{ Pal20041887,basu2005handwritten,basu2009hierarchical,das2009handwritten,Das2010,Pal2012, Bag2013}. However, the collection of literature on handwritten \textit{Bangla} compound character recognition is far shallower as the majority of the  literature focuses only on \textit{Bangla} numeral and basic character recognition.

The work on handwritten \textit{Bangla} compound characters is relatively newer and started during the last decade. Handwritten compound characters bring an added variability to the characters and make for a challenging pattern recognition problem. In \cite{Pal4418297}, features extracted from the directional information of the arc tangent of the gradient in combination with a Modified Quadratic Discriminant Function classifier were used for handwritten compound character recognition on a problem size of 138 classes of compound characters. In another work \cite{das2009handwritten}, Quad tree based features were used for recognition of 55 frequently occurred compound characters covering 90\% of the total of compound characters in \textit{Bangla} using an Multilayer Perceptron (MLP) classifier.In another work of the same author \cite{Das2010}, 50 Basic and 43 frequently used Compound characters are considered to create a 93 class character recognition problem which was solved using shadow and quad tree based longest run features and MLP and support vector machine as classifiers. In \cite{Bag5734941}, a method to improve classification performance on \textit{Bangla} Basic characters using topological features derived from the convex shapes of various strokes was proposed. 

More recently, \cite{Das20152054} describes handwritten \textit{Bangla} Character recognition using a soft computing paradigm embedded in a two pass approach. More specifically highly misclassified classes were combined to form a single group in the first pass or coarse classification.  In the second pass, group specific local features were identified using Genetic Algorithm based region selection strategy to classify the  appropriate class from the groups formed in the earlier pass. They used two different sets of features -- a) convex hull based features b) Longest run based features  with Support Vector Machines (SVM), a well known classifier for this purpose. They reported a recognition accuracy of 87.26\% on a dataset of handwritten \textit{Bangla} characters consisting of Basic characters, Compound characters and Modifiers. In another work \cite{sarkhel2016multi}, Sarkhel et al approached the issue from a perspective of multi-objective based region selection problem where the most informative regions of character samples were used to train  SVM classifiers for character recognition. Two algorithms for optimization, specifically a Non-dominated Sorting Harmony Search Algorithm and Non-dominated Sorting Genetic Algorithm II were used to select the most informative regions with the objective of minimal recognition cost and maximal recognition accuracy. A recognition accuracy of 86.65\% on a mixed dataset of \textit{Bangla} numerals, basic and compound characters was reported. In Section \ref{sec:prev}, we discuss another methodology on handwritten \textit{Bangla} compound character recognition on CMATERdb 3.1.3.3, a standard freely available handwritten \textit{Bangla} compound character dataset. 

\subsection{Description of CMATERdb 3.1.3.3: Isolated Handwritten \textit{Bangla} Compound Character Database}
\label{sec:dataset}
The CMATERdb 3.1.3.3  is a database having 171 unique classes of isolated grayscale images of \textit{Bangla} compound characters. There are 34439 individual training samples and 8520 test samples in the dataset. The images in the dataset, being neither centered nor of uniform scale, constitute a difficult pattern recognition problem.

For ease of comparison, Table \ref{table:diffmnistcmater} presents the fundamental characteristics of the widely used MNIST dataset and the CMATERdb 3.1.3.3 dataset. It is apparent that a higher number of classes, unnormalized images in terms of scale and translation, and with a lower amount of training data, the basic pattern recognition problem with CMATERdb 3.1.3.3 is more complicated than that in the MNIST dataset.

\subsection{Methodology of previous benchmark}
\label{sec:prev}
The previous benchmark \cite{Das2014} (which happens to be the original benchmark) on the CMATERdb 3.1.3.3 isolated compound character database used the traditional methodology of feature engineering in combination with a classifier. In particular, a feature descriptor composed of convex hull and Quad Tree based features was used with an SVM classifier. 

For the Quad Tree based features, a quad tree of depth 2 was created where the partitioning of the image was done at the centre of gravity of the black pixels. This recursive partitioning resulted in 21 sub-images representing the nodes of the Quad Tree. From these sub-images, row-wise, column-wise and diagonal-wise longest run features \cite{Basu2009postal} are extracted. For the convex hull based features, multiple 'bay' and 'lake' descriptors are defined based on the convex hull around the black character pixels \cite{Das2010}. The combined features are used to train a Support Vector Machine for character recognition.

An error rate of approximately 19\% was achieved as the erstwhile benchmark on the 171 class dataset. 

\section{Proposed Recognition Strategy}

\subsection{Deep Convolutional Neural Network}
The simplest description of a DCNN would be to imagine the LeNet-5 Convolutional Neural Network \cite{LeCun1998} with an arbitrarily high number of Convolution and Pooling layers. The architecture of a DCNN consists of possibly alternate Convolution or Pooling layers, before ending up in one or many fully connected layers and an output layer. Concepts of shared weights, local receptive fields and subsampling that are essential to a DCNN have been discussed repeatedly in literature \cite{LeCun1989,LeCun1998,Ciresan2011,Bengio2009} and have not been repeated here. 

\subsection{Supervised Layerwise trained Deep Convolutional Neural Networks}

\begin{figure}[!t]
	\centering
	\includegraphics[height=8.5cm,width=9cm]{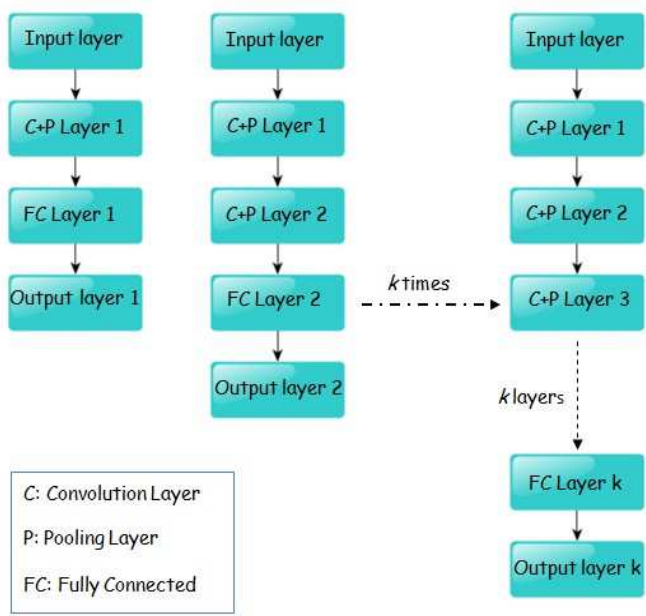}
	\caption{Block diagram of an SL-DCNN layered k times}
	\label{fig:SL-DCNN}
\end{figure}

\begin{algorithm}[!t]
	\SetAlgoLined
	\SetKwFunction{AddLayer}{AddLayer} \SetKwFunction{RemoveLayer}{RemoveLayer} \SetKwFunction{TrainModel}{TrainModel}
	\KwData{trn, tst, lrate1, lrate2}
	\KwResult{A trained SL-DCNN model}
	\BlankLine
	\Begin{
		$trn \longleftarrow \mbox{Training Data} $\;
		$tst \longleftarrow \mbox{Testing Data} $\;
		$INPUT \longleftarrow \mbox{Layer specifying dimensions of image} $\;
		$CP \longleftarrow \mbox{Convolutional and Pooling layer} $\;
		$FC \longleftarrow \mbox{Fully Connected layer} $\;
		$OUTPUT \longleftarrow \mbox{Specifies the number of output classes} $\;
		$lrate1 \longleftarrow \mbox{Comparatively High Learning Rate} $\;
		$lrate2 \longleftarrow \mbox{Low Learning Rate} $\;
		\BlankLine
		\tcp{model is initialized as null}
		$mdl \longleftarrow \phi$ \;
		\BlankLine
		\AddLayer{mdl,INPUT}\;
		\AddLayer{mdl,CP}\;
		\AddLayer{mdl,FC}\;
		\AddLayer{mdl,OUTPUT}\;
		\tcp{RMSProp is used when training model}
		\TrainModel{mdl, trn, tst, lrate1}\;
		\BlankLine
		\tcp{index here points to last layer}
		$index \longleftarrow -1$\;
		\BlankLine
		\While{not all \emph{layers} have been added}{
			\RemoveLayer(mdl, $index$)\;
			\RemoveLayer(mdl, $index$)\;
			\AddLayer{mdl,CP}\;
			\AddLayer{mdl,FC}\;
			\AddLayer{mdl,OUTPUT}\;
			\TrainModel{mdl, trn, tst, lrate1}\;			
		}
		\BlankLine
		
		\TrainModel{mdl, trn, tst, lrate2}\;
	}
	\caption{\rule{0pt}{2.5ex}Technique for training an SL-DCNN}
	\label{algo:layerwiseDCNN}
\end{algorithm}

For supervised learning using any particular DCNN architecture, to get the most out of the model, we propose an alternative to straight forward training on the entire model architecture. Greedy Unsupervised Layerwise \cite{Bengio2007} training as applied to models like denoising autoencoders suggest training the neural network layer by layer and then a fine tuning of the network. A supervised variant of the unsupervised layerwise training methods, made popular in literature in the last decade, is put forward as a superior training method for DCNNs without any significant change to model architecture or training algorithms. 

Supervised layerwise training of a Deep Convolutional Neural Network (SL-DCNN) essentially involves training a DCNN by building the model incrementally by adding Convolution and Pooling layers one by one and training them before adding more layers. After the entire model is built, a fine tuning is performed by training the entire model at a very low learning rate for a short number of iterations. In fact, our method has the same 2 steps that are common to unsupervised layerwise training \cite{Vincent2010} – that is, layerwise training and fine tuning. The only difference is that the cost function for unsupervised training has been replaced with a cost function for supervised training based on classification error. Figure \ref{fig:SL-DCNN} illustrates an SL-DCNN layered \textit{k} times.

Since there is no obvious way of reusing the output layer and the fully connected layer, we choose to retrain a new of each of the above with every new layer added. The reason for doing so is that the use of at least one fully connected layer seems to stabilize the learning process when added. Hence, with the computational cost of adding the 2 layers being minimal, a fully connected layer is added every time a new convolutional and pooling layer is added. The old fully connected and output layers are obviously removed before the addition of new ones.

In (deep) convolutional neural networks, the relatively low fan-in of individual neurons coupled with the use of Rectified Linear activation units do not allow the gradient to diffuse much. However, issues such as convergence to local minima as well as lower generalization performance do persist and layerwise training has been shown to be beneficial (\cite{Erhan2010}) when applied to such problems. The same is observed when supervised layerwise training is applied by us to DCNN training.

While there are no architectural differences between the DCNN and SL-DCNN, the method of realization and training of the complete model is where the contrast lies. Layerwise training enables an SL-DCNN to be have an iterative increase in model complexity. However, in case of a DCNN, it involves training the complete model at once. The training methodology of an SL-DCNN demonstrates distinctly faster convergence and higher generalization as compared to the comparatively simple DCNN.

\subsection{RMSProp}
The concept of local learning rates were introduced to avoid a flat global learning rate and enable faster training and better convergence -- all of which are desirable in deep learning systems.

RMSProp is a technique conceptualized by Hinton et al \cite{rmsproplecture} for adaptive local learning rates. It is essentially, a customized version for mini-batch gradient of another adaptive learning rate method called rprop \cite{Riedmiller1993}, which was designed for stochastic gradient descent.

The method works by keeping a moving average of the squared gradient for each weight and dividing the gradient by the square root of this value. RMSProp has been recently discussed in literature \cite{Dauphin2015,carlson2015stochastic} and has been shown to significantly hasten convergence despite possessing certain limitations. The learning rate adjustment parameter at time (or iteration $t$) for each weight $r_t$ is given by,

\begin{equation}
\label{eqn:RMSPropadjust}
r_t = (1-\gamma) f'(\theta_t)^2 + \gamma r_{t-1}
\end{equation}

where $\gamma$ is the decay rate, $f'(\theta_t)$ is the derivative of the error with respect to the weight at time (or iteration) $t$ and  and $r_{t-1}$ is the previous value of the adjustment parameter.

The update to the weights is then given by,
	
\begin{equation}
\label{eqn:updateRMSProp}
\theta_{t+1} = \theta_t - \frac{\alpha}{\sqrt{r_t}} f'(\theta_t)
\end{equation}		

where $\alpha$ is the global learning rate, and $\theta_t$ and $\theta_{t+1}$ are the values of the weight at time (or iteration) $t$ and $t+1$ respectively.

\subsection{SL-DCNN with RMSProp}
RMSProp has been used generously with SL-DCNN thus speeding up layerwise training as well as fine tuning. Unlike unsupervised layerwise models,  supervised layerwise models have no way of utilizing vast amounts of unlabeled data. However, it maybe said without exaggeration that an SL-DCNN model combined with RMSProp represents the most efficient way of utilizing labeled data. Algorithm \ref{algo:layerwiseDCNN} demonstrates the method for building an SL-DCNN layer by layer. 

\begin{figure}[!t]
	\centering
	\includegraphics[width=8.8cm, height=7cm]{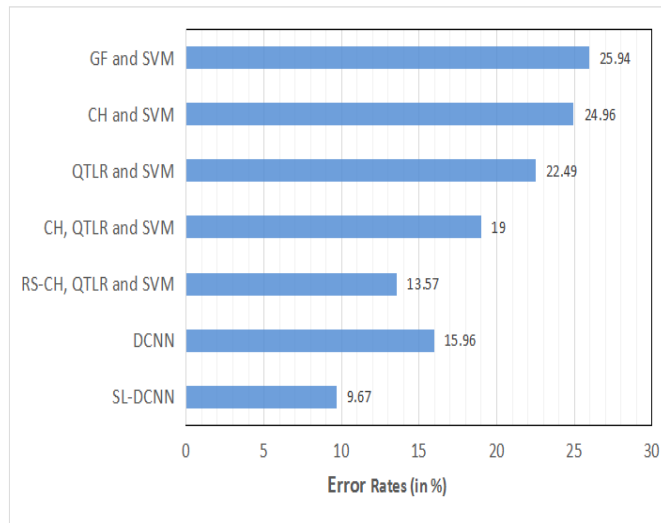}
	\caption{Error rates of the different models on the CMATERdb 3.1.3.3. CH, QTLR and SVM represents the old benchmark on the dataset.}
	\label{fig:barError}
\end{figure}

\section{Experiments}
A series of experiments were performed to compare the convergence rates of the SL-DCNN and the DCNN, as well as to establish a new benchmark on the CMATERdb 3.1.3.3. The experiments were run on a system having an Intel Core i3 processor, 4GB RAM, with Ubuntu 14.04 operating system. The models were coded in Python and used the Pylearn2 machine learning framework \cite{pylearn2_arxiv_2013} and dependencies such as Theano \cite{2016arXiv160502688short}, NumPy and SciPy. Pillow (fork of PIL) was used for basic image processing tasks while HDF5 (h5py) was used for handling dataset storage issues.

\subsection{Preprocessing}
Due to the nature of the dataset and the learning model used, the preprocessing on the dataset was kept to a minimum. The data was subjected to standardization only and resized to a width and height of 70 pixels to make it convenient for the DCNN.

\begin{figure}[!t]
	\centering
	\includegraphics[width = 8.6cm, height=7cm]{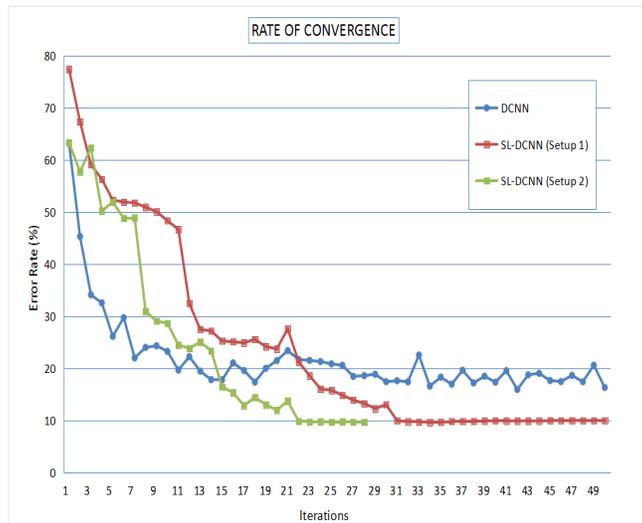}
	\caption{Rates of Convergence of DCNN and SL-DCNN (Setup 1 and 2). Setup 2 error rates are nearly equivalent to Setup 1 error rates}
	\label{fig:Converge}
\end{figure}

\subsection{Description of DCNN and SL-DCNN architectures}
Owing to the difficulty of hyper-parameter search in DNNs, a uniform architecture was used for both types of models, with a standard 4x4 filter size in all layers. The fully connected layer in all models consist of 1500 units. The convention used throughout the article to describe architectures is elaborated below.

\begin{itemize}
	\item \textbf{xCy:} A convolution layer with $x$ number of filters and a filter size of $y$X$y$.
	\item \textbf{xPy:} A pooling layer with a pool size of $x$X$x$ and a pool stride of $y$.
	\item \textbf{xFC:} A fully connected layer of $x$ number of hidden units.
	\item \textbf{xSM:} A softmax output layer of $x$ number of output units.
\end{itemize}
Based on the above convention the architecture used can be written as 64C4-4P2-64C4-4P2-64C4-4P2-1500FC-171SM. It is worth mentioning that for regular DCNNs, the architecture is trained by taking all layers at once, while the SL-DCNN is trained in a layerwise fashion.

\begin{figure*}[!t]
	\centering
	\includegraphics[width = \columnwidth, height=7cm]{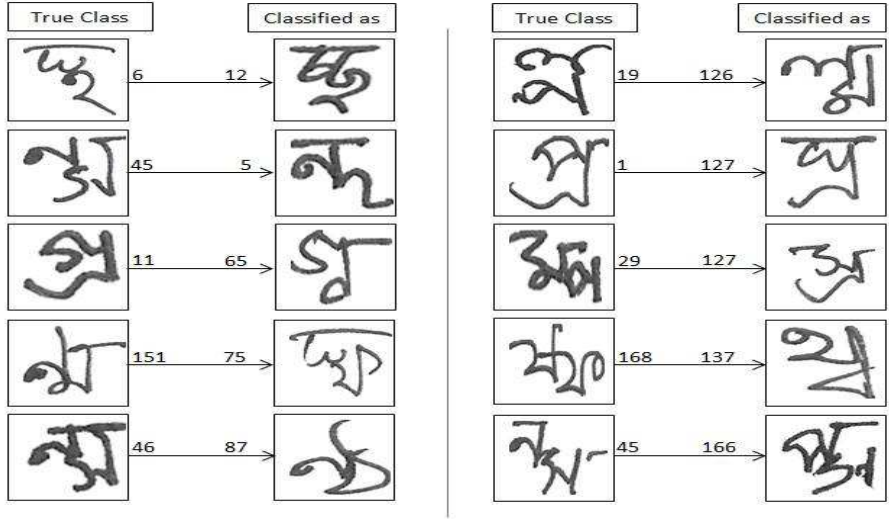}
	\caption{A sample of frequently misclassified classes by the final SL-DCNN model}
	\label{fig:misclass}
\end{figure*}

\subsection{Learning Process Details}
The techniques, mentioned below, were kept consistent across the various DNN experiments to aid the learning process.

\begin{enumerate}
	\item Mini-batch gradient descent: Mini-batch gradient descent with a batch size of 100 was used for all our experiments. 
	\item Rectified Linear Units: Rectified Linear Activation Units were used as the activation units for the neurons in all neural network models. 
	\item Softmax Output Units: Softmax output units were used in the output layers of all the models.
	\item Decaying Learning Rate: In our experiments, the learning rate was made to decay from the original value to 0.1 times of the original value during a total of 50 iterations.
	\item Shuffled Datasets: Shuffling of training examples has been shown to benefit training of neural networks (\cite{LeCun2012}) and was used in all our experiments.  
	\item RMSProp: The RMSProp algorithm as discussed earlier was used in all experiments performed as it significantly expedited the learning.
\end{enumerate}

\subsection{Experimental Setup}
Two different experimental setups are designed by  using different number of iterations per layer for the SL-DCNN for analytical purposes.

\begin{enumerate}
	\item \textbf{Setup 1:} Each Convolution and Pooling layer of the SL-DCNN is trained for 10 iterations, and after addition of 3 Convolution and Pooling layers, a fine tuning is performed for 20 iterations at a very low learning rate. This results in a total of 50 iterations -- 30 for 3 Convolution and Pooling layers and 20 for the fine tuning. This is the same as the number of iterations used by the DCNN. This setup is used primarily to establish a benchmark on the CMATERdb 3.1.3.3 by placing both the DCNN and SL-DCNN on an equal footing.
	\label{intext:Setup1}
	
	\item \textbf{Setup 2:} Each Convolution and Pooling layer of the SL-DCNN is trained for 7 iterations, and after addition of 3 Convolution and Pooling layers, a fine tuning is performed for 7 iterations. This leads to an SL-DCNN model trained to nearly half as many iterations (to be precise 28) as the DCNN model. The purpose of this setup is to illustrate the rate of convergence of the SL-DCNN.
	\label{intext:Setup2} 
\end{enumerate}

\section{Results and Analysis}

\subsection{Benchmark on the CMATERdb 3.1.3.3.}
One of the major aims of this work was to achieve a new benchmark on the CMATERdb 3.1.3.3. As described in Section \ref{sec:prev}, the previous benchmark on the dataset was around 19\% error rate (CH, QTLR and SVM -- \cite{Das2014}). Additionally, we also provide reference results to alternative shallow learning based techniques, each used to train an SVM classifier --- Quad Tree Longest Run features (QTLR and SVM -- \cite{das2009handwritten, wen2007handwritten}), Gradient features (GF and SVM -- \cite{basu2009hierarchical}), Convex Hull features (CH and SVM -- \cite{Das2010}), Region Sampled Convex Hull and Quad Tree Longest Run features (RS-CH,QTLR and SVM - \cite{sarkhel2016multi})  --- for classifying the CMATERdb 3.1.3.3 isolated compound character dataset. We also compare with a regular DCNN model with similar architecture. Our present work surpasses the previous benchmark with our SL-DCNN model providing an error rate of 9.67\% compared to 15.96\% for the regular DCNN.

Figure \ref{fig:barError} illustrates the final error rates on the various models mentioned previously, with the DCNN and SL-DCNN models following parameters mentioned in Setup 1.

\subsection{Rate of Convergence}
The rate of convergence is a compelling argument as to why SL-DCNN as a model represents a highly efficient deployment of a convolutional neural network. With no change in architecture and a change in the training methodology, significant gains are achieved using the SL-DCNN. To make this self-evident, 2 different experimental setups are employed for the SL-DCNN as previously stated.

Figure \ref{fig:Converge} illustrates the rate of convergence for DCNN and SL-DCNN (Setup 1 and Setup 2) . Setup 1 demonstrates that the SL-DCNN converges faster and to a higher accuracy than a DCNN. We have also observed that with the usage of RMSProp as a component for the layerwise training, the models can converge to equivalent accuracies at lower number of iterations. To illustrate this, we have used Setup 2.

Setup 2 is however designed to make the SL-DCNN run for nearly half the total number of iterations as in Setup 1. It was seen in our experiments with Setup 2, that an error rate of 9.69\% was achieved which is nearly equivalent to the error rate of 9.67\% in Setup 1. The model in Setup 2 was seen to converge at iteration 25. This demonstrates that the SL-DCNN trained at each layer to a very low number of iterations still gives a comparatively low error rate. Hence, it can be said that the SL-DCNN model possesses a fast rate of convergence.

\subsection{Analysis of Misclassification}

Although providing a significant reduction in error rates, there are still difficulties inherent to the dataset that the benchmark model struggled with. Inspection of the structure of the characters, along with an analysis of misclassifications reveals some of the issues which makes \textit{Bangla} compound character recognition a particularly challenging pattern recognition problem. Figure \ref{fig:misclass} represents some samples of frequent misclassifications by the final SL-DCNN model. Two reasons for such misclassifications are apparent:

\begin{enumerate}
	\item As mentioned previously, there exist some \textit{Bangla} coumpound character classes with having high structural similarities. One type of such similarities are those found in pairs of compound characters such as 45 and 5 or 46 and 87 (shown in Figure \ref{fig:misclass}), which share one common basic character for true class and misclassified class, hence implicitly showing structural similarity.
	
	\item Also, there are pairs such as 168 and 137 or 6 and 12 (also shown in Figure \ref{fig:misclass}) which, despite not sharing any basic \textit{Bangla} character, are written in a manner where they share major structural constructs.
\end{enumerate}

\section{Conclusion}
DCNNs in general, regardless of the hardware available, are costly to train but extremely effective machine learning models. Hence, fast convergence should be viewed as an important front of research. Having said that, SL-DCNNs represents such a model with higher generalization and faster convergence compared to regular DCNNs.

As a machine learning model, SL-DCNN shows promise for use in allied areas of research and the exploration of the same remains an issue that maybe explored in future works. Also, there remains the possibility to augment the ability of SL-DCNNs by leveraging recent advancements in the area of deep learning such as Dropout, Maxout units, \textit{significantly deeper} networks and Transfer Learning. 

However, the particular SL-DCNN architecture used in this work has been powerful enough to set a new benchmark of 9.67\% error rate on the rather difficult CMATERdb 3.1.3.3 handwritten \textit{Bangla} isolated compound character dataset. This represents a lowering of error rates by nearly 10\% from previously set benchmarks on the same and is an excellent result in the area of \textit{Bangla} compound character recognition.

\section*{Acknowledgments}
This work has been supported by the \textit{Center for Microprocessor Applications for Training, Education and Research (CMATER)} laboratory at the Department of Computer Science \& Engineering of Jadavpur University, Kolkata. The resources provided by the lab have been invaluable in helping us progress with our research.

\bibliographystyle{acm}
\bibliography{refs}


\end{document}